Title: CARE-SD: Classifier-based analysis for recognizing and eliminating stigmatizing and doubt marker labels in electronic health records: model development and validation

Corresponding Author:
Drew Walker
Decatur GA
andrew.walker@emory.edu

Authors:

Drew Walker, (1 and 2), Annie Thorne (3), Sudeshna Das (4), Jennifer Love (5), Hannah LF Cooper, (1), Melvin Livingston III (1), Abeed Sarker (4 and 6)

Affiliations:
1. Department of Behavioral, Social, Health Education Sciences, Rollins School of Public Health, Emory University, Atlanta GA, USA
2. Department of Pediatrics, College of Medicine, University of Florida, Gainesville, FL, USA
3. Department of Infectious Disease, Children's Healthcare of Atlanta, Atlanta GA, USA
4. Department of Biomedical Informatics, School of Medicine, Emory University, Atlanta GA, USA
5. Department of Emergency Medicine, Mount Sinai, New York, NY, USA
6. Department of Biomedical Engineering, Georgia Institute of Technology and Emory University, Atlanta, GA, USA

Dr. Walker's work was supported by NIDA T32 Grant (T32 DA0505552) and Drs Das and Sarker's work was supported by NIDA R01 Grant Funding (R01DA057599)

**Abstract**

**Objective**: To detect and classify features of stigmatizing and biased language in intensive care electronic health records (EHRs) using natural language processing techniques. **Materials and Methods**: We first created a lexicon and regular expression lists from literature-driven stem words for linguistic features of stigmatizing patient labels, doubt markers, and scare quotes within EHRs. The lexicon was further extended using Word2Vec and GPT 3.5, and refined through human evaluation. These lexicons were used to search for matches across 18 million sentences from the de-identified Medical Information Mart for Intensive Care-III (MIMIC-III) dataset. For each linguistic bias feature, 1000 sentence matches were sampled, labeled by expert clinical and public health annotators, and used to supervised learning classifiers. **Results**: Lexicon development from expanded literature stem-word lists resulted in a doubt marker lexicon containing 58 expressions, and a stigmatizing labels lexicon containing 127 expressions. Classifiers for doubt markers and stigmatizing labels had the highest performance, with macro $F_1$-scores of .84 and .79, positive-label recall and precision values ranging from .71 to .86, and accuracies aligning closely with human annotator agreement (.87). **Discussion**: This study demonstrated the feasibility of supervised classifiers in automatically identifying stigmatizing labels and doubt markers in medical text, and identified trends in stigmatizing language use in an EHR setting. Additional labeled data may help improve lower scare quote model performance. **Conclusions:** Classifiers developed in this study showed high model performance and can be applied to identify patterns and target interventions to reduce stigmatizing labels and doubt markers in healthcare systems.

## BACKGROUND AND SIGNIFICANCE

### Provider biases and stigmatization of patients drive healthcare inequities

Provider stigmas and biases are widely believed to contribute to discrimination and health inequities among patients.[1,2] Patients regularly experience stigma and biases by providers as a result of their race, gender, sexual orientation, disease status, drug use, and socioeconomic status, or other labeled characteristics identified by provider teams.[3] Current approaches to reducing provider stigmatization often struggle with limited effect over time, with difficulties in establishing ongoing accountability to ensure long-term behavioral change.[3] The ability to identify potential instances of patient stigmatization within the electronic health record (EHR) could help to inform and target future interventions and facilitate real-time audits of healthcare team communication.

### Stigma defined

Stigma has been defined by social psychologists Link and Phelan as a social process that is characterized by the interplay of *labeling*, *stereotyping* and *separation*, which leads to status loss and discrimination, and importantly, occurs within a context of power, such as that of the patient-provider relationship.[4] Borrowing from Link and Phelan, linguistic bias has been defined as "a systematic asymmetry in word choice as a function of the social category to which the target belongs".[5] Although there are a variety of ways in which linguistic bias can manifest,[6] they all facilitate the transmission of "essentialist beliefs about social categories", resulting in the separation and stigmatization of others.

**Linguistic manifestations of stigmatization**

Beukeboom and Burgers recently developed a framework to understand how stigmatization can manifest and be formed through language, via the Social Categories and Stereotype Communication Framework.[7] This framework posits that stereotypes are communicated through systematic differences in how other groups are 1) labeled , and 2) how their behaviors and characteristics are described. Recent research investigating linguistic bias in the EHR has identified salient features of bias within the language of provider notes, namely focused on the usage of stigmatizing labels, negative descriptors, as well as the expression of doubt in patient testimony through linguistic doubt markers and “scare quotes”.[8–10]

Stigmatizing labels

Stigmatizing labels to describe groups are often used to perpetuate stereotypes, and when used by providers, can lead to feelings of stigmatization and reduced trust among their patients. A recent NIDA study published a list of words to avoid using around patients with substance use disorders, including "addict", "abuser", "user", or "junkie", which have been found to be associated with perceived stigmatization by patients. [11] Similar studies have been applied to other chronic illness populations, identifying terms like “sickler” or “frequent flier” which may be used to further stigmatize patients with chronic illnesses who are often admitted into the hospital. [9,12] While some providers may argue that these terms may be useful in flagging unwanted patient behaviors or mental states, a recent study has shown that patients exposed to language written about them by providers which included stigmatizing labels resulted in patients feeling unfairly judged, labeled, and disrespected.[13] Recent research led by Michael Sun and colleagues on over 40,000 clinical notes has found disparities in presence of “Negative Descriptor” words, which included commonly used terms in the EHR such as “(non-)adherent, aggressive, agitated, angry, challenging, combative, (non-)compliant, confront, (non-)cooperative, defensive, exaggerate, hysterical, (un-)pleasant, refuse, and resist”. Research into stereotype expression in language has found that even seemingly innocuous category labels may prompt others to perceive target individual actions and characteristics as “static” aspects of their identity, and exaggerate differences across groups and similarities within them.[5]

Doubt Markers

Linguistic features such as evidentials, defined as “the linguistic coding of epistemology”,[14] are words that are frequently used in chart language to question the veracity of patients. Among the many words used as evidentials, words and expressions used to confer doubt or uncertainty such as: *allegedly*, *apparently*, or verbs like *claimed*, are often used to when describing patient testimonies, for example: “patient *claimed* their pain was 10/10”.[15] Providers may use words when describing patient testimony in combination with stigmatizing labels or negative descriptors of patients to transmit their stance, or expression of attitudes, feelings, and judgment about patients to other providers which may impact future treatment and care decisions.[16] Inequities have been found among usage of these terms across race and gender, where patients who were women and patients who were Black were found to have significantly higher frequencies of evidentials in their provider notes than patients who were men or White.[15]

Scare Quotes

Another linguistic marker of uncertainty that has been previously identified in patient charts are “scare quotes”, which involve the utilization of quotation marks to mock, cast doubt, challenge patient credibility, or insinuate low health literacy when describing the testimony of another individual.[15] While quotations in charts can be useful to describe patient symptoms using their exact language and document patient wishes or concerns, recent linguistic research has identified a troubling prevalence of providers utilizing quotations in ways to mock, manipulate, and regulate the voices of patients. For example, consider the ambiguity added to the sentence: "Patient reports 10/10 pain related to sickle cell crisis.", when you add "Patient reports '10/10' pain related to 'sickle cell crisis'. Because of the quotation marks, both 10/10 and sickle cell crisis could be inferred as being untrue or uncertain. Similar to evidentials and negative patient descriptors, scare quotes have been found to be more prevalent among patients who were Black and among patients who were women.[8,15]

**Natural Language Processing to identify stigma and bias in EHR Data**

While linguistic studies have acted to guide researchers into identifying manifestations of stigma and bias in text, due to the nature of qualitative, in-depth assessments, these methods can have significant limitations in being able to be deployed to rapidly identify stigma and bias in medical notes in a way in which we can intervene on it. Recent advances in computational linguistics are allowing researchers to harness human-annotated insights on linguistic bias and stigma and scale up categorization to larger amounts of unlabeled data.

This paper aims to apply advanced methods in natural language processing to detect and assess the presence of stigmatizing and biased language in the EHR for patients within the ICU. This study extends qualitative and NLP methods to detect provider stigma bias in EHR in an effort to increase the accountability, evaluation, and intervention of negative bias that threatens to deteriorate quality of care for patients among a variety of uniquely stigmatized groups. Building from other research in linguistic stigma and bias in medical charts, this study represents the development of the most comprehensive, automated classification system for doubt markers and stigmatizing labels, applied on the largest corpus of de-identified EHRs to date.[17] Doubt marker and stigmatizing label lexicons, as well as classification models, are available for others to utilize this pipeline to identify stigmatizing language in EHR. [18]

## METHODS

Our methods for identifying sentences within the MIMIC-III EHR dataset containing doubt markers, stigmatizing labels, and scare quotes within provider notes consisted of three steps:

1) Lexicon development and sample preparation
2) Sentence-level annotation, and
3) Supervised classification using bag-of-words and transformer-based models

**MIMIC-III Dataset**

The Medical Information Mart for Intensive Care, or "MIMIC-III", is a freely-available database of comprehensive, de-identified EHR, free-text notes, and event documentation for over 40,000 patients admitted to the ICU at Beth Israel Deaconess Medical Center in Boston, MA from 2001 to 2012.[19] This dataset contains over 1.2 million clinical provider notes, across nearly 50,000 admissions. Because this dataset contains freely-available, EHR from ICU providers from a diverse range of conditions and age ranges, it is a valuable resource for developing bias and stigma detection algorithms in provider language, particularly for patients living with chronic illnesses who may be more likely to be admitted to critical care units.[20]

## Lexicon development and sample preparation

The lexicon development process for doubt markers and stigmatizing patient labels began with a stem-word list describing words previously identified as demarcating doubt or perpetuating stigmatizing patient labels within medical charts. We expanded expanded these word lists to include misspellings or words with high semantic similarity and relevance in the domain by using two subsequent techniques: 1) BioWordVec, a word embeddings model trained on medical text, which generated the top 10 most semantically similar words for each stem word,[21,22] and 2) GPT 3.5, which suggested an additional 25 words and spelling deviations for each lexicon, following chain-of-thought prompting related to each linguistic bias feature.[23] Following the first round of expansion, we manually validated the list of generated words for task relevance, and assessed human annotation interrater reliability on whether each word was relevant to each specific bias feature. After the second round of GPT 3.5 expansions, we assessed 10-20 sample matches from the highest top-frequency terms' to remove any extremely high-frequency word matches from the lexicon which were not related to transmission of stigmatizing labels or doubt markers and could have significant impact on the annotation sample. This iterative process, reliant on expert-driven inquiry, and complemented by unsupervised, supervised, and transfer learning methods, reflects the strategies championed by computational grounded theory framework, and ensures our results are informed and validated by human domain experts.[24] Our analytic pipeline is outlined in Figure 1, with intermediary results described in Appendix 1.

**Figure 1: Natural language processing analytic pipeline for lexicon development, regular expression matching, annotation, and classifier model training for stigmatizing linguistic features in MIMIC-III.**

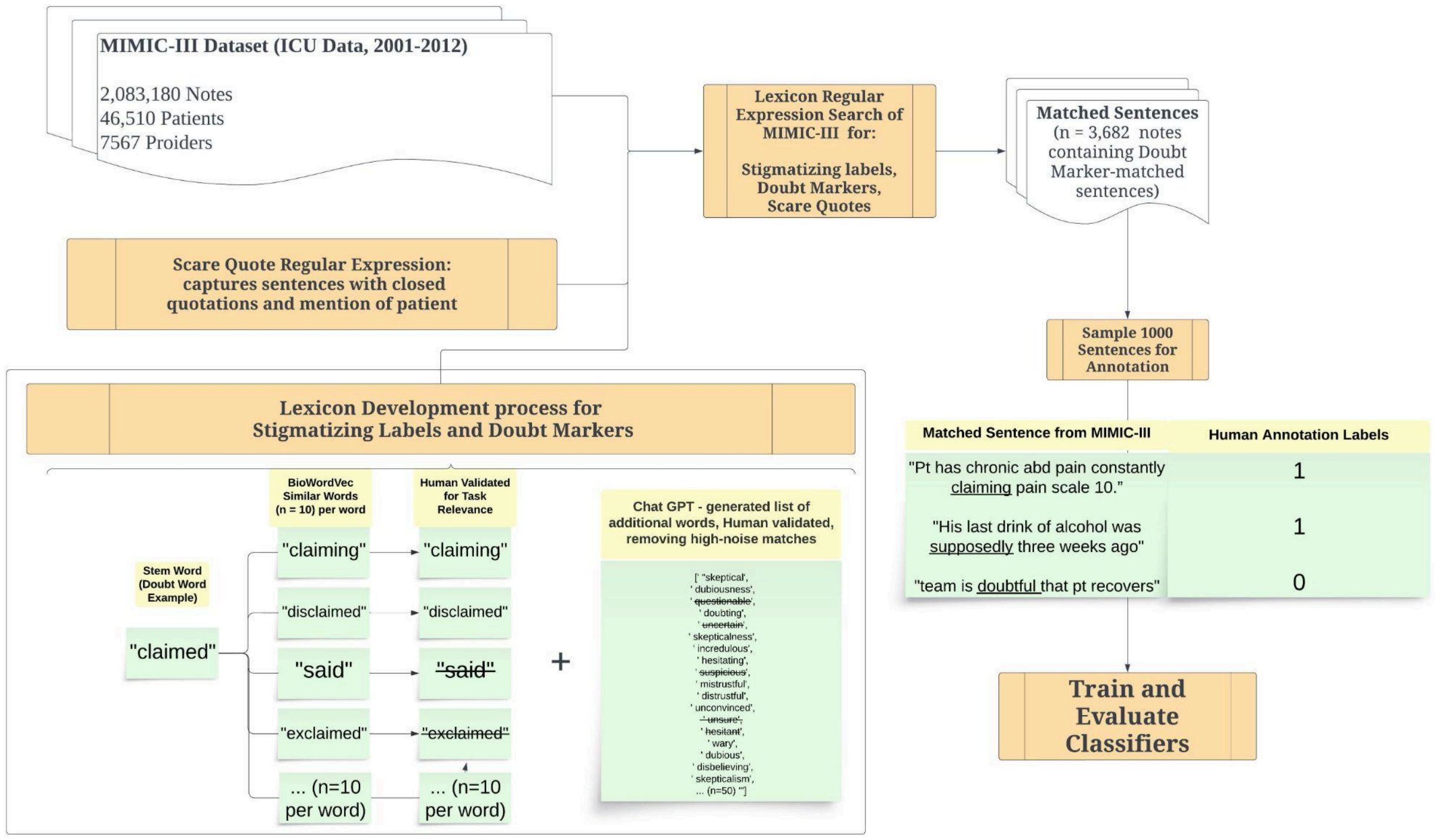

### Stigmatizing labels

Stigmatizing label lexicon development was guided by literature on stigmatizing language in medical care, specifically from the NIDA “Words Matter” publication, Sun’s “Negative Patient Descriptors: Documenting Racial Bias in the Electronic Health Record”, as well as Zestcott’s “Health Care Providers’ Negative Implicit Attitudes and Stereotypes of American Indians”. [11,25,26] The initial stem word list consisted of 18 words: "abuser","junkie","alcoholic", "drunk", "drug-seeking","nonadherent", "agitated", "angry", "combative", "noncompliant", "confront", "noncooperative", "defensive", "hysterical", "unpleasant", "refuse","frequent-flyer", "reluctant".

### Doubt markers

Doubt marker lexicon development was guided by literature on use of “doubt markers” in medical care, specifically led by Beach and colleagues, which identified words such as “claims”, “insists”, and “adamant” or “apparently”, which have been found to be used to discredit or invalidate patient testimony. The 6 words included on the initial stem list were: "adamant", "claimed", "insists", "allegedly","disbelieves","dubious".

### Scare quotes

Scare quote sample preparation was created by searching the MIMIC-III notes using regular expression ((?=.*\".*\")(?=.*\b(pt|patient|pateint|he|she|they)\b)) which caught matching closed quotes, and references to patients by “patient” derivations and pronouns, in order to more accurately capture quotes with patient attributions. Finally, several words were added to filter rows, where matches with quoted words were commonly referring to answers for “alert and oriented” examinations -- i.e. “Patient Name”, “Hospital”, “Year”, etc.

### Matching with sentences in MIMIC-III, creating coding samples

Expanded lexicons or regular expression for each linguistic bias feature were used to filter through patient free-text clinical notes, which had been tokenized at the sentence level to allow for easier readability and classification feasibility. All duplicate sentences were removed from the dataset, and charts labeled as EEG or Radiology were removed in order to restrict to charts more likely to have subjective narrative and patient history text data, such as progress notes, history and prognosis notes, and discharge summaries. Each linguistic feature dataset was randomly sampled in non-replacement groups of 100 (for double-coder reliability scores), 400 (coded by AT, a Physician’s Assistant), and 500 (coded by DW, a behavioral data scientist).

## Annotation process

Coding ontologies for each of the 3 linguistic bias features were developed originally by DW, then iterated on during the first round of reliability coding. The original ontologies were inspired by research led by Beach, Park, and Goddu on the role of stigmatizing language in patient charts.[9,15,27] Qualitative annotators met once to discuss each of the three coding ontologies and guiding theories, as well as co-code 5 sentence examples from each linguistic bias dataset. Following the first meeting, each coder completed the same set of 100 sentences for each of the linguistic bias feature datasets. After inter-rater reliability was assessed, the coders met to discuss disagreements and sentences marked as “close calls”, or difficult labeling decisions, and “exemplary” sentences, which were particularly obvious examples to review. After all disagreements were adjudicated by the coders, they solo-coded 400 (AT) and 500 (DW) sentences to complete the 1000 samples for each linguistic bias feature.

**Sentence Classification**

The annotation data was used to train supervised models for the binary classification task of identifying sentences which do, or do not contain each of the linguistic bias features. This supervised learning task was carried out using four models: Naive Bayes, Logistic Regression, Random Forest, as well as the state-of-the-art transformer-based RoBERTa model.[28] Sentences from clinical notes were tokenized into 1-2 word unigrams and bigrams

A grid search approach was used for hyperparameter optimization for these models, using the training data set, which was split at 80/20%. For Naive Bayes, Logistic Regression, and Random Forest, we utilized a stratified k-fold, with 5 splits, in order to create training and test sets which preserved the percentage of samples for each class.

Following model training, each model was evaluated on a held-out 20% of the data, in which we prioritized the performance metrics of 1) positive-class precision, 2) positive-class recall, and 3) macro $F_1$-score to select the highest-performing model of each model type. Hyperparameter values for each of the best-performing models for each linguistic feature are available in Appendix 2-4. Positive-class precision, or the proportion of true positives divided by the total number of positive predictions, was prioritized to reduce false positives and develop models highly likely to identify actual stigmatizing language. Recall, or the proportion of all true positives labeled correctly, was prioritized to ensure our classifier can identify as many true cases of stigmatization as possible. $F_1$ scores, or the harmonic means of recall and precision, provide a measure of balance between the two performance metrics for each model across both positive and negative class labels. We applied bootstrapping to model evaluation by assessing prediction and ground-truth labels of 1000 samples, generated without replacement. The performance metrics of $F_1$, precision, accuracy, recall were aggregated to calculate the confidence intervals of all model metrics.

Finally, we assessed best-performing model text feature importance and feature logistic regression coefficients in order to evaluate the degree to which certain matched terms and phrases contribute to the linguistic bias label predictions of the models. Feature importance was assessed using Gini importance mean impurity reduction method for decision trees in the random forest classifiers, and regression coefficients were calculated from the logistic regression classifiers.[29–31] Code for all analyses for this study are available on GitHub: https://github.com/drew-walkerr/Diss_Detecting_Provider_Bias.

## RESULTS

**Lexicon Development**

For the stigmatizing labels lexicon expansion and annotator pruning, the initial list of 18 was expanded to 180, which was then assessed by annotators DW and SD, removing 83 terms (Annotator agreement = 75%). Final decisions were adjudicated by DW. The final expanded and pruned list of stigmatizing labels used to search the MIMIC-III dataset totaled 127 words, and is provided in Appendix 1. Following assessments of most frequent term matches, we removed the following terms due to high proportion of noise referring to illness characteristics or clinical situations, rather than patients or patient testimonies: 'difficult', 'suspicious','aggressive','unstable', 'dramatic', 'unreliable','entitled','invalid','violent', 'dangerous'.

For the doubt markers lexicon expansion and annotator pruning, the initial list of 6 terms was expanded to 60, which was then pruned by annotators to remove 2 terms (Annotator agreement = 80%). The final expanded list of doubt markers used to search the MIMIC-III dataset totaled 58 words, and is provided in Appendix 1. Following regular expression searching and assessments of most frequent term matches, we removed the following terms due to high proportion of noise referring to uncertainty in illness or clinical presentations, rather than patient testimonies: 'suspicion', 'suspicious', 'questionable', 'questioning', 'uncertain', 'hesitancy', 'hesitant','unsure'.

**Figure 1: Network Diagram showing doubt marker stem words (green, also includes GPT 3.5) to show expansions of initial stem words to expanded lexicon words (purple)**

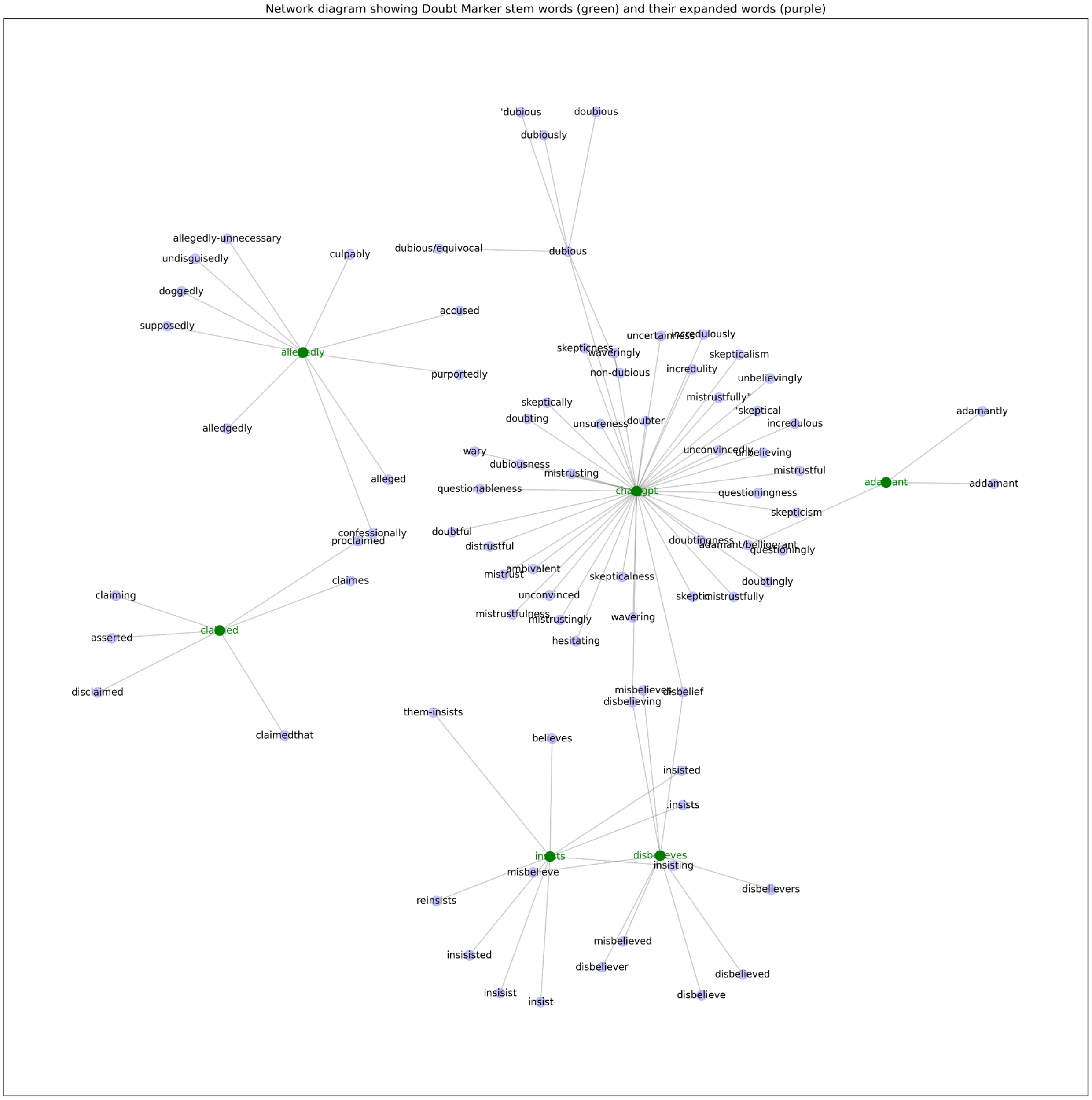

**Regular Expression Search results**

Results describing the text data of the preprocessed MIMIC-III full sample, as well as of search results for each of the stigmatizing labels, doubt markers, and scare quotes regular expression (regex) matched data frames are summarized in Table 1.

**Table 1: Summary statistics of MIMIC-III Dataset, compared with linguistic bias corpa**

| | MIMIC-III Sample | Stigmatizing Label Corpus | Doubt Marker Corpus | Scare Quotes Corpus |
|---|---|---|---|---|
| Number of unique notes | 814,548 notes | 8,950 notes | 3,682 notes | 4,806 notes |
| Avg note length | 654 words | 623 words | 937 words | 763 words |
| Number of total sentences | 18,288,213 sentences | 10,278 sentences | 3,856 sentences | 5,156 sentences |
| Average sentence length | 12 words | 48 words | 35 words | 55 words |
| Number of patients | 11,633 patients | 3,483 patients | 2,368 patients | 2,830 patients |
| Number of providers | 1,879 providers | 1,056 providers | 800 providers | 677 providers |

The most frequent matching terms from our lexicon, along with the most commonly occurring trigrams within quoted text, are provided for each of the 3 bias features in Figure 2. For stigmatizing labels, versions of ‘refusing’ and ‘refuses’ were by far the most frequently matched terms. In the doubt marker label lexicon, ‘believes’ was the most frequently matched term, followed by insisted and insisting. Scare quote quoted text frequent words, bigrams, and trigrams were less led by any particular phrases or words, but were mostly used describing patient chief complaints, descriptions of symptoms or condition.

**Figure 2: Top 20 Matched Terms Stigmatizing Labels, Doubt Markers, and Scare Quotes**

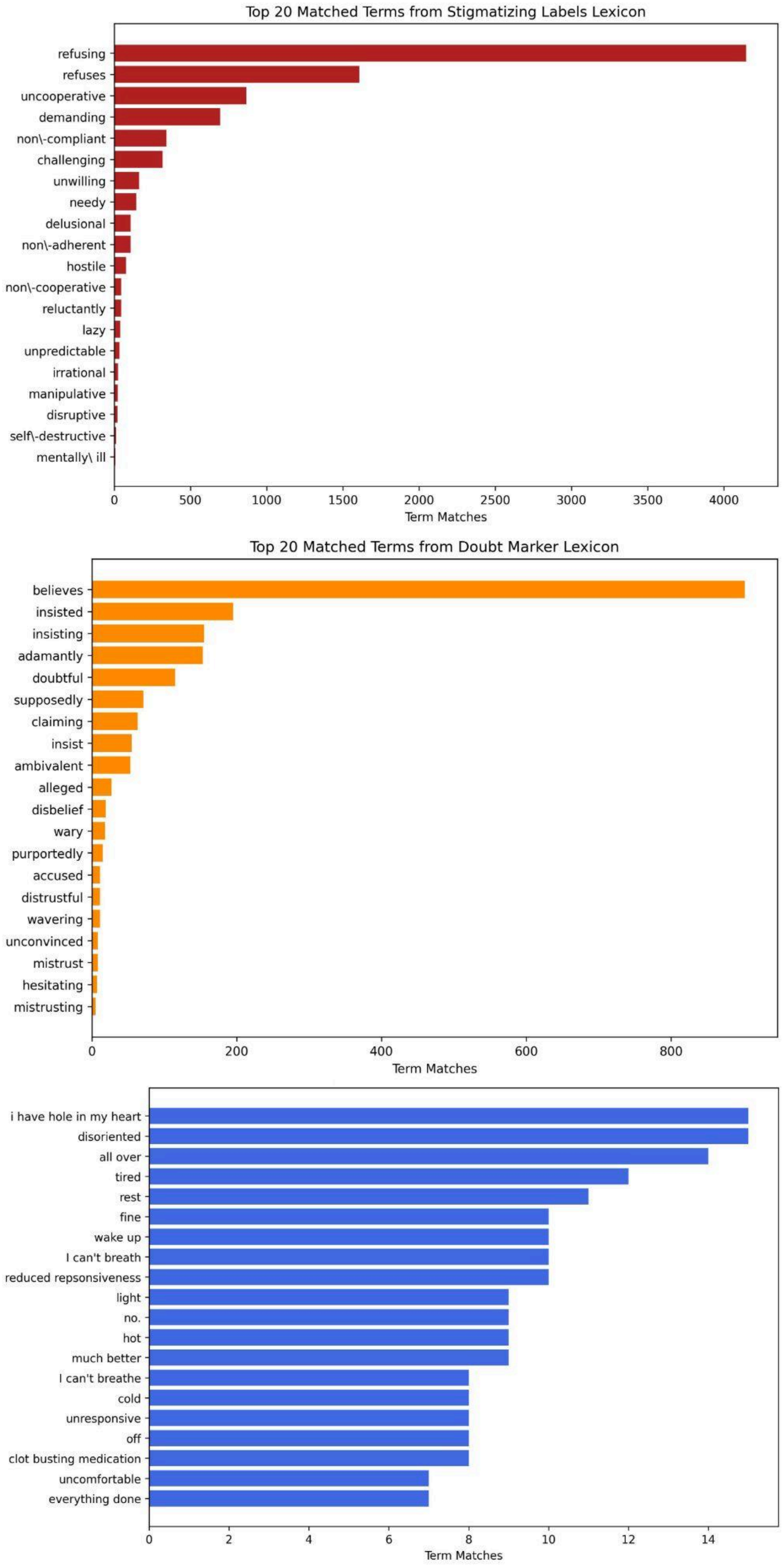

**Annotation**

Annotation coding ontologies, detailing the labeling instructions for each of the three linguistic bias features, were informed largely through the literature-based stem word operationalizations of each set of linguistic bias features. Appendices 1-3 detail the coding ontologies for each corpus. Table 2 provides the interrater agreement and kappa score for the first 100 samples of each linguistic feature, and positive class frequency in the final 1000 sentence sample.

**Table 2: Annotation sample reliability, linguistic bias features positive class frequencies, and notable examples**

| **Bias Feature** | **Agreement** | **Interrater reliability** (Kappa) | **Frequency in final sample** (N=1000) |
|---|---|---|---|
| **Stigmatizing Labels** | 87% | .74 | 43.9% |
| **Doubt Markers** | 87% | .73 | 31.0% |
| **Scare Quotes** | 87% | .73 | 20.7% |

Table 3 provides notable positive class examples for each of the 3 linguistic features.

**Table 3: Notable annotation examples for stigmatizing labels, doubt markers, and scare quotes**

| **Bias Feature** | **Notable Sentence Examples**<br>(Flagged as containing biased feature, matching or quoted text underlined) |
|---|---|
| **Stigmatizing Labels** | Pt very uncooperative, will barely allow any nursing care.<br><br>Neuro: Is very needy, needs to be incouraged to do more for herself.<br><br>Refuses blood draws, calling out, asking for dilaudid despite med not being due, requiring much emotional support. |
| **Doubt Markers** | “Pain control (acute pain, chronic pain) Assessment: Pt has chronic abd pain constantly claiming pain scale 10.”<br><br>“His last drink of alcohol was supposedly three weeks ago.”<br><br>Insisting on making phone calls becomes very irritated if don't jump to his requests. |
| **Scare Quotes** | Easily frustrated, especially when asked questions to assess orientation...pt states "you have already asked me this 100 times".<br><br>NO nausea/vomiting although pt states he does not want to eat because he "fears" being nauseated.<br><br>At 10:06, pt put on call light to request "pain pill" and then put on the call light 5 more times over the next 8 minutes to make same demand, at one point saying to the nurse responding to the call light, "What the hell is going on." |

### Linguistic Bias Classifier Model Evaluation Results

Table 4 displays the results of the best performing models across types and linguistic bias features. A complete list of the best performing model hyperparameters can be found in Appendix 5. We were able to achieve the highest performance on doubt markers and stigmatizing labels, with scare quote models underperforming other linguistic bias models across nearly every evaluation metric.

The best performing model type for doubt markers and scare quotes was RoBERTa, with Logistic Regression achieving the best performance in classifying stigmatizing label sentences.

Run duration was much higher for RoBERTa models, as compared with Random Forest, Logistic Regression, and Naive Bayes classifiers.

**Table 4: Linguistic Bias Classifier Model Performance (best model of each feature in bold)**

| Bias Feature | Model | Accuracy | Precision (Positive) | Recall (Positive) | F1 (Positive) | Macro Precision | Macro Recall | Macro F1 | Run Duration |
|---|---|---|---|---|---|---|---|---|---|
| **Stigmatizing Labels** | RoBERTa | .69 (.63, .76) | .63 (.54, .72) | .75 (.67, .84) | .69 (.62, .76) | .70 (.64, .76) | .70 (.64, .76) | .69 (.63, .75) | 484.1s |
| | Random Forest | .79 (.73, .84) | .77 (.68, .86) | .73 (.63, .82) | .75 (.67, .82) | .77 (.68, .86) | .73 (.63, .82) | .75 (.67, .82) | 148.4s |
| | **Logistic Regression** | **.81 (.75, .86)** | **.75 (.66, .85)** | **.84 (.75, .91)** | **.79 (.72, .86)** | **.75 (,66, .84)** | **.84 (.75, .91)** | **.79 (.72, .86)** | **3.2s** |
| | Naive Bayes | .71 (.64, .77) | .62 (.53, .70) | .86 (.79, .93) | .72 (.64, .78) | .62 (.53, .70) | .86 (.79, .93) | .72 (.64, .78) | 0.1s |
| **Doubt Markers** | **RoBERTa** | **.86 (.81, .91)** | **.86 (.75, .96)** | **.71 (.59, .81)** | **.77 (.69, .85)** | **.86 (.80, .91)** | **.82 (.76, .88)** | **.84 (.78, .88)** | **790.72s** |
| | Random Forest | .85 (.80, .89) | .76 (.64, .86) | .76 (.65, .85) | .76 (.67, .84) | .76 (.64, .86) | .76 (.65, .85) | .76 (.67, .84) | 18.0s |
| | Logistic Regression | .85 (.80, .90) | .71 (.61, .81) | .89 (.80, .96) | .78 (.69, .85) | .70 (.60, .80) | .88 (.80, .96) | .78 (.69, .85) | 2.4s |
| | Naive Bayes | .85 (.80, .89) | .70 (.60, .80) | .89 (.80, .96) | .78 (.69, .85) | .70 (.60, .80) | .89 (.80, .96) | .78 (.69, .85) | 0.1s |
| **Scare Quotes** | **RoBERTa** | **.75 (.69, .81)** | **.40 (.24, .58)** | **.30 (.17, .45)** | **.35 (.20, .48)** | **.61 (.52, .70)** | **.59 (.52, .67)** | **.62 (.52, .70)** | **810.0s** |
| | Random Forest | .79 (.74, .85) | 0.00 (0,0) | 0.00 (0,0) | 0.00 (0,0) | 0.00 (0,0) | 0.00 (0,0) | 0.00 (0,0) | 148.9s |
| | Logistic Regression | .77 (.71, .82) | .30 (.07, .56) | .10 (.02, .20) | .14 (.04, .28) | .30 (.07, .56) | .10 (.02, .20) | .14 (.04, .28) | 1.7s |
| | Naive Bayes | .78 (.72, .83) | .43 (.22,.63) | .24 (.12, .38) | .31 (.16, .45) | .43 (.22, .63) | .24 (.12, .38) | .31 (.16, .45) | 0.1s |

After model evaluation, we also ran feature importance and contribution plots using the models of the best performing random forest and logistic regression models. Figure 3 highlights terms or phrases that are particularly informative to random forest models during categorization (left-hand side), and the right-hand side displays the terms with the highest regression coefficients (negative and positive), which are more likely to be labeled as negative (blue, unbiased), or positive (red, biased/stigmatizing).

**Figure 3: Top 30 Stigmatizing Label Tokens by Importance and Feature Contributions**

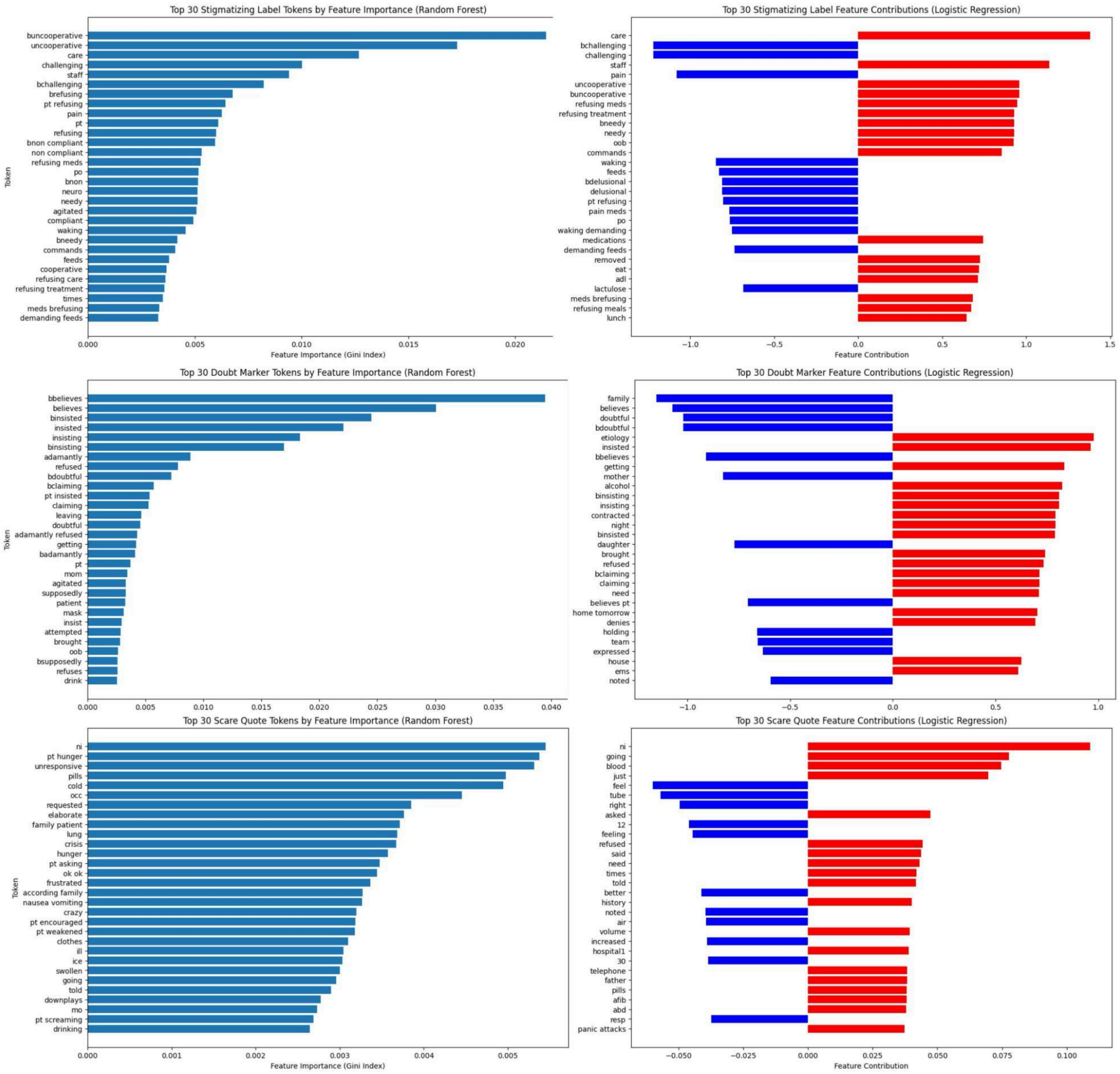

# DISCUSSION

This study demonstrated the viability of scaling up previous research on stigma in language to develop tools that can be applied to identifying stigmatizing text in the EHR. Our suite of tools, comprising lexicons and refined supervised classification models, titled "CARE-SD 1.0: Classifier-based Analysis for Recognizing and Eliminating Stigmatizing and Doubt Marker Labels in Electronic Health Records" are available on GitHub for other researchers in this space seeking to reproduce or adapt our study process.

For the task of lexicon development, our expanded word lists successfully produced terms that frequently matched with sentences which were eventually labeled as stigmatizing or doubt-marking by coders. These lexicons are valuable tools for researchers seeking to focus on EHR sentences with a higher frequency of stigmatizing or doubt marking signals. This process can be adapted to expand searches for words related to a variety of concepts within stigma and other sociolinguistic phenomena.

Supervised learning classifiers for stigmatizing labels and doubt markers demonstrated near-human agreement performance. We believe that these tools, trained on ICU data among a large cohort of patients, will be applicable to other EHR settings or datasets since the linguistic markers are likely to be semantically similar.

Several patterns in stigmatizing label and doubt marker use arose that may help inform clinical practice. Among stigmatizing labels, the use of "needy" was predictive of positive stigmatizing labels. "Needy" was often used to describe an inherent trait, which is highly problematic for anyone seeking care in the ICU. Other providers may wrongfully assume the patient is "needy" in other contexts, including pain management, daily activity assistance, or any other genuine health complaint."Noncompliant", a word identified in the expanded stigmatizing labels lexicon, was labeled as stigmatizing chiefly when it was used to define the patient directly as such and not when describing a specific behavior or when providing more context (i.e. due to lack of funds). Labeling patients as "noncompliant" has been hotly debated, though recent EHR-related NLP work is increasingly operationalizing its use as one that has negative connotations, and points blame at the patient, rather than structural factors which may have stronger impacts and constraints on patient health.[17,25]

We encountered many usages of "refusing" among stigmatizing label charts. We aimed to negatively label (unbiased) when providers were specific about the behavior and context in which the patient was refusing a specific treatment, or when the context was surrounding end-of-life or do not resuscitate (patient refusing further life-saving measures), and aimed to positively label (biased) when the language around what patients were refusing was vague, or painted the patient as refusing any care in a way that could portray the patient as inherently stubborn. We also positively labeled sentences which described patients "refusing to try" or "refusing to perform [activity of daily living]". Our coding ontology posits that depicting a patient as refusing to try/assert effort can be interpreted as labeling a patient as lazy or unwillful, when they may be unable to complete tasks due to discomfort or decreased health status.

Information gathered from doubt markers feature importance and contribution plots, as well as from annotator notes and feedback also show patterns that reflect doubt marker usage across a variety of contexts. Use of "insisted" and "insisting" were consistently labeled as positive for doubt markers. While these words may be helpful in expressing the strong conviction behind patient needs, it is frequently also laden with negative, stubborn, or difficult connotations. The word "claimed", particularly when describing pain or severity of illness symptoms or pain, was highly likely to be labeled as a doubt marker. "Alcohol" also had a high feature contribution towards positive doubt marker labels. Both annotators described that doubt marker words were frequently employed to report patient alcohol and drug use history, as exemplified in Table 2.

**Limitations**

Low performance on scare quotes may be indicative of the need for additional data to train future models. Due to the high variety of quote usage in language, it is important to consider different approaches and linguistic structural features within scare quotes. Additionally, our dataset consists only of notes from ICU admissions from one site. Many stigmatized patients, particularly those with mental health or substance use problems, may not be admitted due to a variety of structural barriers, interactions with healthcare providers, or concomitant illness. [32–34]

Another limitation to consider is the positionality of labelers in this study. Annotations were completed by two researchers, who, although trained in reading medical texts and interpreting linguistic bias, have limited scope to the variety of experiences of stigmatization and bias which many patients may endure.

Finally, it is important to convey that as language and medicine have changed from 2001-2012 in MIMIC-III notes, there may be words that were acceptable at one time and are only realized to be derogatory over time. The process of identifying stigma and bias must be a continuous effort to keep pace with how stigma and biases evolve over time to evade scrutiny or draw in-group members closer together.[35,36]

**Conclusions**

The CARE-SD1.0 models and lexicons produced in this study hold high utility for identifying patterns of stigmatizing labels and doubt markers in healthcare systems, particularly for targeting and designing interventions. Our goal in sharing these tools is to accelerate research efforts in the study of linguistic stigma and bias in healthcare. With additional validation, these models can be used to audit and evaluate healthcare systems for units, providers, or patients whom experience higher rates of stigmatizing labels and doubt markers, and allow real-time feedback for anti-stigmatization intervention efforts in a way that has not been possible with traditional implicit bias training approaches.[3]

## APPENDIX 1: LEXICONS FOR DOUBT MARKERS AND STIGMATIZING LABELS

| Lexicon | Stem Word List | Expanded Words (Pruned to) | GPT-3.5 added words | High-noise terms removed | Final Lexicon | Final Lexicon Length |
|---|---|---|---|---|---|---|
| **Doubt Markers** | "adamant", "claimed", "insists", "allegedly","disbelieves","dubious" | 60 total, reduced by 2. Agreement = 80% | [' "skeptical', ' dubiousness', ' questionable', ' doubting', ' uncertain', ' skepticalness', ' incredulous', ' hesitating', ' suspicious', ' mistrustful', ' distrustful', ' unconvinced', ' unsure', ' hesitant', ' wary', ' dubious', ' disbelieving', ' skepticalism', | 'suspicion', 'suspicious', 'questionable', 'questioning', 'uncertain', 'hesitancy', 'hesitant','unsure' | ['"doubtful', '"dubious", '.insists', 'accused', 'adamant', 'adamant/belligerant', 'adamantly', 'addamant', 'alledgedly', 'alleged', 'allegedly', 'allegedly-unnecessary', 'asserted', 'believes', 'claimed', 'claimedthat', 'claimes', 'claiming', 'confessionally', 'culpably', 'disbelief', 'disbelieve', 'disbelieved', 'disbeliever', 'disbelievers', 'disbelieves', | 58 |
| | | | ' hesitancy',<br>' skepticism',<br>' mistrust',<br>' uncertainness',<br>' disbelief',<br>' suspicion',<br>' mistrustfulness',<br>' incredulity',<br>' incredulously',<br>' wavering',<br>' ambivalent',<br>' waveringly',<br>' questionableness',<br>' mistrustingly',<br>' doubter',<br>' questioning',<br>' doubtingly',<br>' mistrusting',<br>' doubtful',<br>' skeptic',<br>' unconvincedly',<br>' mistrustingly',<br>' mistrustfully',<br>' doubtingness',<br>' skepticism',<br>' questioningness',<br>' unbelieving',<br>' unsureness',<br>' skepticness',<br>' questioningness',<br>' doubtingly',<br>' unbelievingly',<br>' skeptically',<br>' mistrustingly',<br>' mistrustfully',<br>' skeptically',<br>' questioningly',<br>' doubtingly',<br>' skeptically',<br>' mistrustingly',<br>' mistrustfully'"] | | 'disbelieving',<br>'disclaimed', 'doggedly',<br>'doubious', 'doubtful',<br>'dubious',<br>'dubious/equivocal',<br>'dubiously', 'insisist',<br>'insisisted', 'insist',<br>'insisted', 'insisting',<br>'insists', 'misbelieve',<br>'misbelieved',<br>'misbelieves',<br>'mistrustful',<br>'mistrusting',<br>'non-dubious',<br>'proclaimed',<br>'purportedly', 'reinsists',<br>'skeptical', 'speculative',<br>'supposedly',<br>'them-insists',<br>'unconvinced',<br>'undisguisedly',<br>'unreliable', 'unsure"',<br>'wavering'] | |
| **Stigmatizing Labels** | "abuser","junkie","alcoholic", "drunk", "drug-seeking","nonadherent", "agitated", "angry", "combative", "noncompliant", "confront", "noncooperative", "defensive", "hysterical", "unpleasant", "refuse","frequent-flyer", "reluctant" | 180 , reduced by 83. Annotator agreement = .75. | [' "hysterical', ' aggressive', ' drug addict', ' non-compliant', ' lazy', ' attention-seeking', ' manipulative', ' hypochondriac', ' difficult', ' mentally unstable', ' troublemaker', ' irresponsible', ' unpredictable', ' irrational', ' needy', ' demanding', ' disruptive', ' uncooperative', ' unreliable', ' high maintenance', ' attention-seeker', ' dramatic', ' attention-seeking', ' lazy', ' invalid', ' faker', ' irrational', ' hostile', ' aggressive', ' challenging', ' | 'difficult', 'suspicious','aggressive','unstable', 'dramatic', 'unreliable','entitled','invalid','violent', 'dangerous' | ['"hysterical',<br>'"drug-seeking",<br>'"hysterical",<br>'"drug-seeking",<br>'"junkie", '.reluctant',<br>'abuse/abuser',<br>'abused-abuser', 'abuser',<br>"abuser's", 'abusers',<br>'addictive-drug-seeking',<br>'alcoholic', 'angry',<br>'angry-disgusted',<br>'angry/disgusted',<br>'attention-seeker',<br>'attention-seeking',<br>'challenging',<br>'combative',<br>'combatively',<br>'compliant/noncompliant', 'counterdefensive',<br>'deceptive', 'defensive',<br>'defensive/offensive',<br>'delusional',<br>'demanding',<br>'disruptive', 'drug addict', 'drug seeker',<br>'drug-craving/drug-seeki | 127 |
| | | | uncooperative', ' deceptive', ' demanding', ' unreliable', ' high-strung', ' self-destructive', ' unstable', ' manipulative', ' entitled', ' attention-seeking', ' violent', ' drug seeker', ' malingerer', ' faker', ' mentally ill', ' dangerous', ' delusional', ' needy', ' overly sensitive', ' unstable', ' irrational'"] | | ng', 'drug-seeking', 'drug-seeking/-taking', 'drug-seeking/drug-taking', 'drug-seeking/taking', 'drug-seeking/use', 'drunk', 'drunken', 'drunkenly', 'drunker', 'drunkest', 'drunks', 'ex-abuser', 'ex-alcoholic', 'faker', 'frequent-flier', 'frequent-flyer', 'frequent-flyers', 'frequent-fvl', 'frequent-hitter', 'frequent-hitters', 'high maintenance', 'high-strung', 'histrionic-hysterical', 'hostile', 'hypochondriac', 'hypochondriac-hysterical', 'hysteric', 'hysterical', 'hysterical-obsessive', 'hysterical/anaclitic', 'hystericals', 'hysterics', 'incompliant', 'irrational', 'irrational'", 'irresponsible', 'iv-abuser', 'ivdabuser', 'junkie', "junkie's", 'junkies', 'lazy', 'ma-abuser', 'malingerer', 'manipulative', 'mentally ill', 'mentally unstable', 'morereluctant', 'needy', 'non-adherent', 'non-alcoholic/alcoholic', 'non-compliant', 'non-cooperating', 'non-cooperation', 'non-cooperative', 'non-cooperatively', 'nonadhered', 'nonadherent', 'nonadherently', 'nonadherents', 'noncompliant', 'noncompliant/compliant', 'noncompliant\\medically', 'noncompliants', 'noncooperating', 'noncooperation', 'noncooperative', 'noncooperatively', 'novelty/drug-seeking', 'ny-nonadherent', 'onadherent', 'overdefensive', 'overly sensitive', 'prealcoholic', 'pt.noncompliant', 'refuse', 'refuses', 'refusing', 'reluctanly', 'reluctant', 'reluctantly', 'reluctants', 'schizo-hysterical', | |

<table>
<tr><td></td><td></td><td></td><td></td><td></td><td>'self-destructive', 'troublemaker', 'un-adherent', 'unadherent', 'uncooperative', 'unpleasant', 'unpleasant/annoying', 'unpleasantly', 'unpleasantries', 'unpredictable', 'unwilling', 'unwillingly']</td><td></td></tr>
<tr><td>Scare Quotes</td><td>((?=.*\".*\")(?=.*\b(pt|patient|pateint|he|she|they)\b))</td><td>-</td><td>-</td><td>['yes','no','itchy','hospital','Do Not Resuscitate', 'Yes or No', 'wet','yes/no','dry', 'DO NOT RESUSCITATE',' comfort measures only', 'Known', 'Name', 'firstname', 'lastname','[**Doctor Last Name **]', '[**Last Name (un) **]', '[**Known firstname **] [**Known lastname **]','[**Doctor First Name **]', '[**Hospital1 **]','[**Hospital3 **]','[**Known lastname **], [**Known firstname **]']</td><td></td><td></td></tr>
</table>

## APPENDIX 2: STIGMATIZING LABELS ONTOLOGY

### *Coding Process*

DW and AT met for 1 hour before the first round of coding began. During this time they discussed rationale and literature backgrounds for each of the three linguistic bias features, then proceeded to co-code 5 examples not included in the subsequent datasets. Then, each annotator coded the same random sample of 100. DW and AT met to discuss each of the disagreements in this sample, and used these examples to further inform ontology development for the final sample. Coders were able to reach agreement on all of the linguistic bias features after discussing disagreements.

After calculating agreement and meeting to adjudicate disagreements, AT coded an additional 400 sentences, and DW coded 500 sentences for each linguistic bias term. Table 2 displays the results of the reliability dataset, including frequencies of class labels, as well as final results from the 1000 sentence dataset, with notable sentence notes, chosen out of a selection marked by coders for potential manuscript examples.

### *Link and Phelan Stigma Definition*

Stigma has been defined by social psychologists Link and Phelan as a social process that is characterized by the interplay of

1. **Labeling**: Identifying individuals as belonging to a particular group. Status loss and discrimination involves a negative evaluation of a groups' attributes and its members relative to another group. Commonly used with nouns (static label) or direct adjectives to patient (patient is insistent on receiving water vs patient was repeatedly asking for water). Nouns/direct adjectives work to label patients as static qualities rather than specific, isolated behaviors.
2. **Stereotyping**: ascribing a proclivity towards a specific behavior or characteristic to members of a labeled group.
3. **Separation**: involves distancing from the group, and drawing lines of "us versus them".
4. **Status loss and discrimination**: involve a negative evaluation of a groups' attributes and its members relative to another group. (Comparing this patient with others, or patient vs provider)
5. **Occurs within a context of power**, such as that of the patient-provider relationship.

(Link & Phelan, 2001)

### *Stigmatizing labels and negative descriptors in charts*

Stigmatizing labels to describe groups are often used to perpetuate stereotypes, and when used by providers, can lead to feelings of stigmatization and reduced trust among their patients. Much of the recent work on identifying and reducing stigmatizing labels has come from providers seeking to improve care for patients with substance use disorders. A recent NIDA study published a list of words to avoid using around patients with substance use disorders, including "addict", "abuser", "user", or "junkie", which have been found to be associated with perceived stigmatization by patients. (Abuse, 2021a) Similar studies have been applied to other chronic illness populations, identifying terms like "sickler", "frequent flier" or "drug-seeking", which may be used to further stigmatize patients with chronic illnesses who are often admitted into the hospital. (Abuse, 2021b; Glassberg et al., 2013; Goddu et al., 2018) While some providers may argue that these terms may be useful in flagging unwanted patient behaviors or mental states, a recent study has shown that patients exposed to language written about them by providers which included stigmatizing labels resulted in patients feeling unfairly judged, labeled, and disrespected.(Fernández et al., 2021)

Recent research led by Michael Sun and colleagues on over 40,000 clinical notes has found disparities in presence of "Negative Descriptor" words, evaluated by the Health Equity Commission of the Society of General Internal Medicine, which included commonly used terms in the EHR such as "(non-)adherent, aggressive, agitated, angry, challenging, combative, (non-)compliant, confront, (non-)cooperative, defensive, exaggerate, hysterical, (un-)pleasant, refuse, and resist". This study found that compared to White patients, Black patients had 2.54 times the odds of having at least one negative descriptor written in their history and physical notes. (Sun et al., 2022) **Research into stereotype expression in language has found that even seemingly innocuous category labels may prompt others to perceive target individual actions and characteristics as "static" aspects of their identity, and exaggerate differences across groups and similarities within them.** (Beukeboom & Burgers, 2017) These labels can be used to justify clinical decision-making, withholding of resources, or to confer doubt upon patient testimonies. (Beukeboom, 2014) While current recommendations encourage use of person-centered, neutral language in medical charts, it is important to evaluate the presence of known stigmatizing labels within provider notes to mitigate the transmission of bias in the EHR.

**Guiding Question:** Does this sentence involve language about the patient which could result in the stigmatization or negative labeling of a patient, which could lead to further status loss/discrimination in the context of the patient-provider relationship?

### *Coding Rules*

- **Code = 1** : Yes, sentence involves language that could result in the stigmatization of this patient-- i.e. it involves labeling, stereotyping, separation, which could lead to status loss/discrimination in the context of the patient-provider relationship.
    - Yes, clear example:
        - He was demanding dilaudid on admission
        - Patient very needy this shift
    - Patient refusing care. (Broadly)
    - Refusing: difficult to discern,
        - can be used in stigmatizing way related to daily activities or effort put forth by patient
            - If the quote was patient "refusing" a normal daily activity, we code =1.
                - Example: "Patient refusing to put on socks" , which could paint them in a negative, or "stubborn" light. We should instead say "patient not able to put socks on" or "patient does not want socks on".
                - This is bc we are placing direct blame on patients due to lack of effort, without considering other factors like inability
        - If the quote was describing how the patient was refusing, related to effort on behalf of the patient, code =1.
            - Examples: "Patient refusing to attempt", "patient refusing to try", "patient refusing to cooperate". These may be more related to patient suffering and not getting care they need, and saying the are refusing to put forth effort can be stigmatizing.
- **Code = 0:** No, the sentence does not involve stigmatizing/negative patient descriptors. Stigmatizing word/negative descriptor is not referring to the patient's static characteristics or would not likely result in status loss/discrimination among the medical team.
    - These are typically when words are used to describe a patients' specific behavior vs painting a picture about their character in broad strokes.
        - I.e. if they're refusing, be specific about what they're refusing
    - Examples:
        - "Waking for some feeds but not demanding yet."
        - "O2 sats were unreliable and could not be monitored"
        - "Non-adherent bandage"

- Instead of calling a patient non-adherent to treatment or noncompliant, which would fall under stigmatizing labels/negative pt descriptors
    - Describing patient acute psychosis or “delusion”
    - “Refusing”
        - Refusing a specific medical treatment, without other stigmatizing language or adjectives, would be a 0.
            - Example: “pt refusing to go to CT scan”
        - Refusing DNR status-- well within patient right to refuse, and it is important to clearly understand . These would be labeled 0.
            - “Refusing further care” == 0 when discussed in the context of being do not resuscitate/ do not intubate

## APPENDIX 3: DOUBT MARKERS ONTOLOGY

### *Coding Process*

DW and AT met for 1 hour before the first round of coding began. During this time they discussed rationale and literature backgrounds for each of the three linguistic bias features, then proceeded to co-code 5 examples not included in the subsequent datasets. Then, each annotator coded the same random sample of 100. DW and AT met to discuss each of the disagreements in this sample, and used these examples to further inform ontology development for the final sample. Coders were able to reach agreement on all of the linguistic bias features after discussing disagreements.

After calculating agreement and meeting to adjudicate disagreements, AT coded an additional 400 sentences, and DW coded 500 sentences for each linguistic bias term. Table 2 displays the results of the reliability dataset, including frequencies of class labels, as well as final results from the 1000 sentence dataset, with notable sentence notes, chosen out of a selection marked by coders for potential manuscript examples.

**Doubt markers overview**

Linguistic features such as evidentials, defined as “the linguistic coding of epistemology”,[1] are frequently used along with other words referred to as “doubt markers”, to question the veracity of patient testimonies, particularly related to their symptoms and adherence to treatment.(Park et al., 2021)

Among the many words used as doubt markers, words and expressions used to confer uncertainty such as: *allegedly*, *apparently*, or verbs like *claimed*, are often used when describing patient testimonies, for example: “patient *claimed* their pain was 10/10”.[3] These words are often used to discuss the veracity of patient symptoms and adherence to treatment.

Disparities have been found among usage of these terms across race and gender, where patients who were women and patients who were Black were found to have significantly higher frequencies of doubt markers in their provider notes than patients who were men or White.[3] The extent to which providers use doubt markers is posited to be reflective of the amount of doubt and uncertainty a provider has on patient testimony, and is thus hypothesized to impact trust within the patient-provider relationship, and related outcomes like leaving against medical advice or in pain management strategies.[2,3]

Providers may use words when describing patient testimony in combination with stigmatizing labels or negative descriptors of patients to transmit their stance, or expression of attitudes, feelings, and judgment about patients to other providers which may impact future treatment and care decisions. [4]

**Guiding Question:**

Could this sentence be interpreted by a provider in a way that confers doubt towards the patient's testimony, behavior, or condition?

***Coding Rules***

**doubt_testimony = 1:** Yes, it could be interpreted to cast doubt on patient testimony, behavior, or condition.

a. Examples:
    i. "apparently he was sitting at home on the floor feeling fine when suddenly he felt fatigued all over his body,"
    ii. "the patient insists she gets sick from vaccines."

**doubt_testimony = 0:** No, this sentence would likely not be interpreted to doubt the patients' testimony, behavior, or condition.

b. These may just reflect uncertainty in medical results/plan
    i. Example: "Diagnosis remains unclear at this time"
c. They could also be instances when the chart is discussing secondhand information about a patient, like "family believes patient is depressed". Because testimony is secondhand, this is actually appropriate.
d. Other times, the word may be used to accurately portray a patient's own doubt or subjectivity around a situation.
    i. Example: "patient believes they have no options left"

## APPENDIX 4: SCARE QUOTES ONTOLOGY

### *Coding Process*

DW and AT met for 1 hour before the first round of coding began. During this time they discussed rationale and literature backgrounds for each of the three linguistic bias features, then proceeded to co-code 5 examples not included in the subsequent datasets. Then, each annotator coded the same random sample of 100. DW and AT met to discuss each of the disagreements in this sample, and used these examples to further inform ontology development for the final sample. Coders were able to reach agreement on all of the linguistic bias features after discussing disagreements.

After calculating agreement and meeting to adjudicate disagreements, AT coded an additional 400 sentences, and DW coded 500 sentences for each linguistic bias term. Table 2 displays the results of the reliability dataset, including frequencies of class labels, as well as final results from the 1000 sentence dataset, with notable sentence notes, chosen out of a selection marked by coders for potential manuscript examples.

### Scare Quotes Overview

Another linguistic marker of uncertainty that has been previously identified in patient charts are "scare quotes", which involve the utilization of quotation marks to mock, cast doubt, challenge patient credibility, or insinuate low health literacy when describing the testimony of another individual.(Beach et al., 2021) While use of scare quotes has been documented since the 1950s, some argue that the linguistic phenomenon has been increasing increasing in recent years, both from the rise of "air quoting" gesture in the 80s and 90s, and was commonly employed by Trump prior to and throughout his presidency.(Garber, 2016; Saner, 2017)

While quotations in charts can be useful to describe patient symptoms using their exact language and document patient wishes or concerns, recent linguistic research has identified a troubling prevalence of providers utilizing quotations in ways to mock, manipulate, and regulate the voices of patients. For example, consider the ambiguity added to the sentence: "Patient reports 10/10 pain related to sickle cell crisis.", when you add "Patient reports '10/10' pain related to 'sickle cell crisis'. Because of the quotation marks, both 10/10 and sickle cell crisis could be inferred as being untrue or uncertain. Similar to evidentials and negative patient descriptors, scare quotes have been found to be more prevalent among patients who were Black and among patients who were women.(Beach et al., 2021) Early research on scare quoting in patient charts has recommended that providers utilize quotes only when absolutely necessary to reflect a patient's exact experience, wishes, or concerns, and that even when made in earnest, unnecessary quotation opens patient testimony up to a level of uncertainty or inference to other providers who may question the veracity of patient conditions or experiences.(Beach & Saha, 2021)

**The goal of this annotation task is to determine whether these sentences identified as matching closed quotation strings and including reference to patient could be interpreted**

**by other providers as “scare quotes”.** We will use a binary system 1/0 to determine whether or not the sentence included a scare quote or not.

We are trying to understand if the use of these quotes could be interpreted to mock, cast doubt, challenge patient credibility, insinuate low health literacy or assign other negative labels to patients. In reading each chart, it’s important to ask: did this need to be quoted? Or could it have been written differently, or more definitively?

### *Coding Rules*

<u>scare_quote = “1”, Could be interpreted as Scare Quote.</u>

- **Yes**: Cast **doubt on** patient as providing **reliable testimony.**
  - Examples:
    - Stated “migraine” was due to “stress”. Vs-- Patient stated migraine was brought on by stress.
- **Yes: Convey ridicule, contempt, stigmatization, or frustration** by highlighting **unsophisticated language** or **limited knowledge**
  - Examples:
    - Patient repeatedly asked to “get me out of this fucking place”
    - Does not believe he has prostate cancer because “his bowels are working fine”

<u>Scare_quote = “0”: Not a Scare Quote.</u>

- **Clinical Info**, **Effect on Life**, **Values or Preferences. Descriptive of issue**
  - Examples:
    - Chest pain that “feels like an elephant is on my chest”
    - Reported that “this is the worst headache I’ve had in my life”
    - When discussing treatment goals, she said “if I cannot breathe without a tube, I don’t want to live. I do not want to suffer. I want to make sure that my family are with me at the end.”
- **Weird random quotes bookending the entire narrative**
  - These will be huge. You’ll know when you see them
- **Describing Acute Psychosis**
  - This is a difficult/tricky line, but in the instances where the quotes are used to describe an aspect of a patients’ psychosis, we believe this is medically necessary to communicate, and not a scare quote, despite that it is used to communicate that the patients’ view is unreliable.
  - Examples:
    - [Pt] then transiently sits up saying clearing "what's going on here"
    - Pt able to sleep intermit overnight, however, pt reported having several bad dreams and awoke very disoriented (calling the nurse the "president's dgtr").

## APPENDIX 5: BEST PERFORMING MODEL HYPERPARAMETERS

**Stigma Model hyperparameters:**

- RF: {'max_depth': None, 'min_samples_split': 2, 'n_estimators': 200}
- NB: {'alpha': 1.0}
- Log Reg: {'C': 1.0}
- RoBERTA: {‘Max_len’:128 , ‘Batch_size’: 4, ‘epochs’ : 10 , ‘learning_rate’: 1e-5}

**Doubt Markers Hyperparameters:**

- RF: {'max_depth': None, 'min_samples_split': 2, 'n_estimators': 100}
- NB: {'alpha': 1.0}
- Log Reg: {'C': 1.0}
- RoBERTA: {‘Max_len’:512 , ‘Batch_size’: 4, ‘epochs’ : 10 , ‘learning_rate’: 1e-5}

**Scare Quote Model Hyperparameters:**

- RF: {'max_depth': None, 'min_samples_split': 5, 'n_estimators': 100}
- NB: {'alpha': 1.0}
- Log Reg: {'C': 0.01}
- RoBERTA: {‘Max_len’:512 , ‘Batch_size’: 8, ‘epochs’ : 10 , ‘learning_rate’: 1e-5}